\documentclass[acmtog]{acmart}

\renewcommand\footnotetextcopyrightpermission[1]{} % removes footnote with conference information in first column
\makeatletter
\def\runningfoot{\def\@runningfoot{}}
\def\firstfoot{\def\@firstfoot{}}
\makeatother

\AtBeginDocument{%
  \providecommand\BibTeX{{%
    \normalfont B\kern-0.5em{\scshape i\kern-0.25em b}\kern-0.8em\TeX}}}

\begin{document}

\title{Searching for Apparel Products from Images in the Wild}

\author{Son Tran}
\email{sontran@amazon.com}
\author{Ming Du}
\email{mingdu@amazon.com}
\author{Sampath Chanda}
\email{csampat@amazon.com}
\author{R. Manmatha}
\email{manmatha@amazon.com}
\author{CJ Taylor}
\email{taylorcj@amazon.com} 

\affiliation{%
  \institution{Visual Search and AR Group, Amazon}
  \streetaddress{130 Lytton}
  \city{Palo Alto}
  \state{CA}
  \postcode{94301}
}

\begin{abstract}
In this age of social media, people often look at what others are wearing. In particular, Instagram and Twitter influencers often provide images of themselves wearing different outfits and their followers are often inspired to buy similar clothes. We propose a system to automatically find the closest visually similar clothes in the online Catalog (street-to-shop searching). The problem is challenging since the original images are taken under different pose and lighting conditions. The system initially localizes high-level descriptive regions (top, bottom, wristwear…) using multiple CNN detectors such as YOLO and SSD that are trained specifically for apparel domain. It  then classifies these regions into more specific regions such as t-shirts, tunic or dresses. Finally, a feature embedding learned using a multi-task function is recovered for every item and then compared with corresponding items in the online Catalog database and ranked according to distance. We validate our approach component-wise using benchmark datasets and end-to-end using human evaluation.
\end{abstract}

\keywords{object localization, object recognition, fashion images, deep neural networks}

\maketitle

\section{Introduction}
\begin{figure*}
 \centering \includegraphics[width=0.60\linewidth]{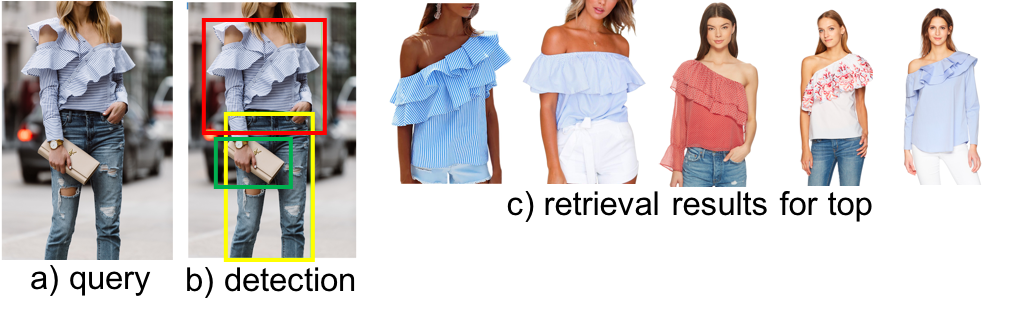}
 \caption{Searching for clothing items from images} 
\label{fig:searching_softlines_items} 
\end{figure*}

People who want to look for fashionable clothes today can look for inspiration to social media, specifically influencers on sites such as Instagram and Twitter. Influencers post images of themselves wearing different outfits in the wild (see Figure \ref{fig:searching_softlines_items}. a). Pose and background are unconstrained and body parts may also be occluded. Items can be layered such as a person wearing a jacket on top of a shirt. All of these lead to challenges in both detecting what people are wearing and also in searching for similar items in the online Catalog. 
%It is also unclear what the person’s notion of similarity means. Are they looking for the same red floral dress or are they looking for a floral dress with a different color? Is the sleeve length more critical or the pattern? Many times, the exact item is not available in the Amazon catalog, so these are difficult questions to answer in the face of approximate results.
%We solve this problem by assuming that people are on a shopping expedition and have a purpose in mind. People often shop by occasion, e.g., for an outfit for the office or for a cocktail party. We must infer this from the query image and provide enough diversity in the results that one or more of them will lead to inspiration and further exploration.

This paper describes our proposed approach to the problem. First, we localize regions of the image at a high-level. We find tops rather than shirts or blouses or jackets and we find bottoms rather than pants or skirts. This allows us flexibility given the large range of possible types of tops or bottoms or the other kinds of regions we find. This localization is done using a YOLOV3 (\cite{Yolov3}) and an SSD (\cite{liu2016ssd}) model trained on apparel categories. We show that how we obtain and train these images does matter by comparing results to those obtained using the OpenImages dataset (\cite{OpenImages}). At this juncture, we also determine gender and coarse age (man, woman, boy or girl) to restrict the corresponding clothes when searching in the database.

Using a hierarchical classifier, we break the results into finer classes – e.g., a \textit{top} may be a \textit{dress shirt} or a \textit{T-shirt} or a \textit{blouse}. We use a CNN with a multi-task loss function that also takes into consideration other attributes, such as \textit{color}, \textit{pattern}, \textit{sleeve length} and so on. The output of the last layer before the softmax is then used as a feature vector for the similarity computation. We performed the same feature extraction on query image and catalog database images. The feature vectors are then compared and ranked. To improve the consistency of our results, the comparisons are restricted to those which are in the same finer class (e.g., \textit{T-shirt} or \textit{blouse}). We present results on the individual components (localization) and also on the overall ranking and compare them with baselines.

The layout of the paper is as follows: related work is reviewed in section \ref{sec:related_work}. Section \ref{sec:system_architecture} describes different major components of our pipeline. The details of the data collection and model training process are in section \ref{sec:data_collection}. Our experimental results are reported in section \ref{sec:experimental} and we provide conclusions in \ref{sec:future}.

\section{Related Work}
\label{sec:related_work}
\subsection*{Localization} We review here a number of main academic approaches for localizing apparel items. In \cite{Deepfashion7780493}, deep CNNs were trained to predict a set of fashion landmarks, such as shoulder points or neck points. However, landmarks for clothes are sometimes not well defined and are often sensitive to occlusions. Human pose estimation (\cite{cao2017realtime})  could be used as the mean to infer apparel item’s locations. One drawback is that they are not applicable if people were not present in the image. Localization could also be carried out through cloth parsing and (semantic) segmentation such as LIP (\cite{Gong_2017_CVPR}). While the performance has been promising on standard datasets, it is more computationally expensive and the recall has been unsatisfactory in the initial evaluation on our targeted datasets (e.g. fashion influencer images). For simplicity and scalability, in this work, we localize apparel items using bounding boxes. Top performing multi-box object detectors such as SSD (\cite{liu2016ssd}), YOLO V3 (\cite{Yolov3}) with different network bodies and different resolutions are used both for run-time queries and offline index construction.

\subsection*{Classification}
Recognizing the product type accurately for apparel is of critical importance in finding similar items from the catalog. In the literature, a number of different classifications have been adopted for clothing items. This is partially a function of what the labeled input datasets provide (\cite{Deepfashion7780493},  \cite{paperdoll}). Most of these papers limit themselves to a small number of classes (less than 60 classes, see, e.g. \cite{zheng2018modanet}) with many high-level classes containing highly dissimilar objects (i.e., different product types). In this work, we create fine-grained classification of 146 classes. Our fine-grained breakdown matches well to product-type levels in online Catalog (for men and women’s clothing) and are typically visually distinctive. Additionally, since we use the Catalog as the search database, matching its hierarchy also greatly facilitates the later indexing and retrieval stages.

\subsection*{Visual Similarity Search}

Visual similarity search can be done by searching for the nearest neighbors to an embedding extracted from certain intermediate layer(s) in a deep neural network trained for surrogate tasks (see \cite{zheng2018modanet}, \cite{FineGrainSimwithDeepWang}, \cite{SunPersonReId} and \cite{NIPS2016_6464}). The deep network can be trained with cross entropy loss (classification), contrastive loss (pairs), triplet loss (\cite{ShankarNAKC17}, \cite{FineGrainSimwithDeepWang}...) or quadruplet loss (\cite{Chen/cvpr2017}). There seem to be no clear winner between these options (see e.g.\cite{SunPersonReId} and \cite{HermansBeyer2017Arxiv}). Obtaining the highest accuracy seems to depend on careful sampling and tuning strategies for the specific problem (see \cite{NIPS2016_6464}). Our approach in this paper, in contrast to parallel efforts in our team, is to directly use the embedding feature extracted from the fine-grained classification network thus avoiding the need for a separate CNN to find an embedding feature. This also helps simplify the engineering efforts and reduce run time latency.

\section{System and Model Architecture}
\label{sec:system_architecture}

In this section, the three main components of our system: localization, recognition, and visual similarity search are first described. This is followed by a description of the overall system and the end-to-end interaction between the components.

\subsection*{Localization}

One could design a detector that detects object/no-object and pass the detected bounding boxes to downstream classifiers for further classification (\cite{girshick14CVPR}). The disadvantage in separating localization of object and its classification in this manner is that spatial context is lost, reducing accuracy. On the other hand, one could try to identify all the fine-grained categories at the detection stage. This would require a very large amount of detailed bounding box annotation to reasonably cover all categories, which is very expensive. (We have found that a 13 high level class division on average takes a well-trained annotator three minutes to complete a bounding box annotation on one image). We strike a middle ground by grouping items of the same types, that often also appear in similar location (e.g. \textit{jacket, coat, t-shirt}) into a high-level class (e.g. \textit{top}). Our complete list of top-level detection class for apparel items is as follows:  \textit{headwear, eyewear, earring, belt, bottom, dress, top, suit, tie, footwear, swimsuit, bag, wristwear, scarf, necklace} and  \textit{one-piece}.

%One key product requirement is getting gender right. We detect gender and coarse age (man, woman, girl and boy) of the apparel items \textit{whether the image under consideration includes people in it or not}. Gender assignment is carried out for each detected non-gender bounding box based on a normalized distance between the gender box and target box.

On the same setting, different object detectors usually have different strengths with respect to object sizes and scales (\cite{Huang2017SpeedAccuracyTF}). To boost recall for offline processes, such as index building, we use an array of multi-box object detectors operating at different levels of image resolutions: SSD-512 Resnet50 (\cite{liu2016ssd}), YOLO V3 Darknet53 300 and 416 (\cite{Yolov3}). Their results are combined using non-maximum suppression. During inference time, we use SSD-512 with a VGG backbone for real-time response.

\subsection*{Fine-Grained Product Type Classification and Feature Extraction}

We build one fine-grained classifier for each of the high-level classes in the previous section. For example, the top classifier will try to classify all detected top bounding boxes into one of the 33 product types such as \textit{denim jacket, tunic, blouse, vest}… %Each of these classes typically corresponds to a leaf node in online Catalog’s browse tree with the proviso that a few visually similar looking nodes such as \textit{pleated shorts} and \textit{flat-front shorts} are merged together into \textit{man shorts}.

Taking run-time into considerations, we chose the Resnet18 (\cite{resnetHeZRS15}) as the backbone for all fine-grained classifiers. To exploit better all available supervised signal in our training sets, we extended the network to perform multi-task classification. For example, for the \textit{top} classifier, we additionally classify \textit{color, pattern, shape, shoulder type, neck type, sleeve type…} For each branch corresponding to one of these tasks, a fully connected layer with a 128-D output is inserted between \textit{pool5} and its \textit{softmax} layers. We observe that by using joint multi-task training, product-type classification as well as search relevance are both improved.
\begin{figure*}
 \centering \includegraphics[width=.60\linewidth]{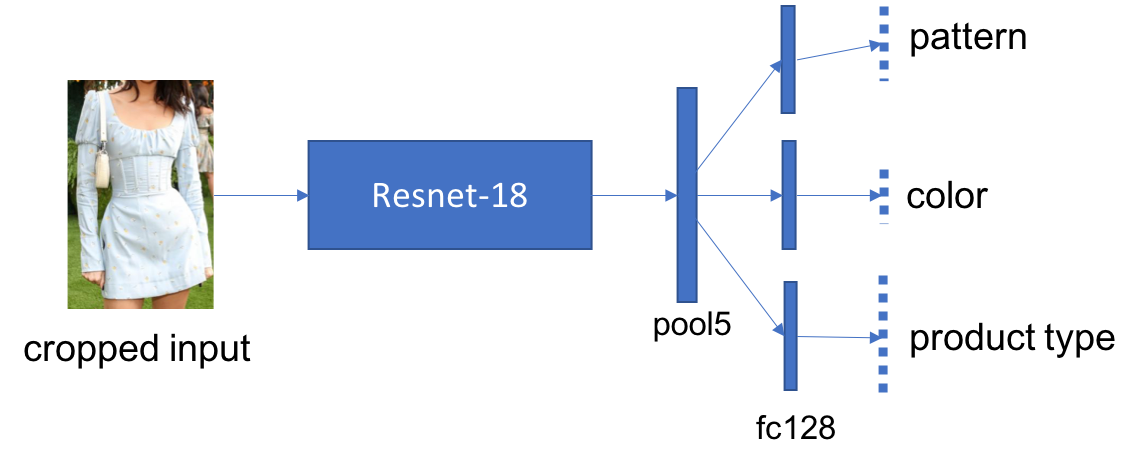}
 \caption{Our fine-grained multitask network for \textit{top}} 
\label{fig:top_network} 
\end{figure*}

The overall loss for the network is calculated as follow: $L =\sum_{i=1}^N L_i*w_i$,  where $N$ is number of tasks, $L_i$ and $w_i$ are the classification loss and its weight for the i-th branch respectively.

The 512-D feature from the \textit{pool5} layer (output of average pooling immediately after all the convolution layers) is used for visual similarity search. With our architecture, illustrated in figure \ref{fig:top_network}, this feature has the power to represent the combined similarity across product type, color, pattern, etc. In addition, the 128-D features from each of the \textit{fc128} layers are specific to each of these aspects and could be used to re-rank the candidate list if so desired.

Previous user research studies show that typically users place the importance of occasion first followed by product type, then color, followed by pattern and other features. Motivated by this observation, the loss weight for product type is empirically set to 1.0, for color, 0.3 and the rest 0.1.
\subsection*{End-to-end System}
\begin{figure*}
 \centering \includegraphics[width=0.75\linewidth]{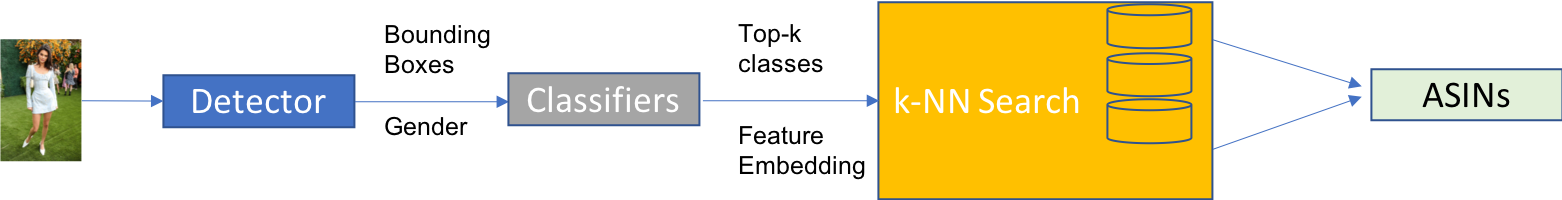}
\caption{Overall architecture of our end-to-end visual search for apparel}
\label{fig:overall_pipeline}
\end{figure*}

The run-time flow of our pipeline is illustrated in Figure \ref{fig:overall_pipeline}. An image is input to the detector which generates high level class bounding boxes (e.g. \textit{top}/\textit{bottom}/\textit{dress}). For each of these bounding boxes, a cropped image patch is extracted and fed into the corresponding fine-grained classifier. The top-k fine-grained classes are identified and their corresponding indices are searched for nearest neighbors (Catalog items) using the embedding feature extracted from the fine-grained classifier. The resulting list of Catalog items from different indices are then re-sorted according to the embedding distance.

\section{Data Collection, Model Training and Implementation Details}
\label{sec:data_collection}
\subsection*{Localization}
We evaluated existing public datasets for apparel and found out that they poorly matched our product needs. For example, OpenImages (\cite{OpenImages}) contains a large amount of bounding box annotations including ones for apparel-related categories. In particular, for shoes, the specific class breakdown and corresponding number of bounding boxes are as follows: (footwear – 744474), (boot – 3132), (high heels – 3124). However, while being large in quantities, they are heavily skewed, incomplete in category, not sufficiently refined and do not match with online Catalog’s partition. Therefore, instead of relying on such public datasets, we decided to download and annotate 50K images from the internet using certain search keyword composition to ensure a balanced distribution across all clothing classes, gender, pose and scenes. A small percentage (5\%) of the resulting images do not contain persons, i.e., only clothing items. We still tried to annotate gender for them to the best we could. Overall, a total of 320K bounding boxes were annotated, an average of 16K boxes per (high level) detection class. This annotation task was carried out by our in-house vendors, since it was observed that using general MTurkers often leads to less accurate bounding box annotation.

We trained the following multi-box detectors on this dataset: SSD-512 VGG, SSD-512 Resnet50, YOLOV3 Darknet53 (300 and 416 image sizes). The first one is used in real-time inference. The remaining detectors were used as an ensemble in bounding box extraction from Catalog images. We used the default training parameters from Caffe (\cite{jia2014caffe}) as well as Gluon-CV (\cite{gluoncv}) libraries and trained each detector for 200 epochs using the vanilla SGD optimizer.

\subsection*{Classification}

Training accurate classifiers requires a large amount of data. Acquiring images from the wild and adding class label for hundreds of apparel categories in a short amount of time is very challenging. Instead, we turned to two available data sources - Catalog and customer review images. The browse node associated with each image is known from the Catalog database. And since our class partition matches with Catalog’s node structure, training class label is readily available at a large scale without the need for manual annotation.

One important issue is duplication, where multiple Catalog items use the same or near identical images. Using such images to train models would result in overfitting. We first applied a filter to eliminate low sale volume items, deduped by image names, and finally deduped using a k-NN engine (similar to deduping for search indices described in the following section).

Another and more critical issue is that Catalog images are typically clean without background clutter, without occlusion or clothing items not being worn on people. This domain difference is well-known and there have been numerous efforts in the literature to close the gap (\cite{Anab}, \cite{WhereToBuyItICCV15}...). In this work, we augment catalog images with random background. Specifically, background patches of random shape and sizes was drawn from a large repository of natural images. Catalog foreground object masks were obtained by first thresholding away white pixels followed by a morphological dilation to create some space around object’s silhouettes. The foreground object is then blended with a random background patch using Poisson image editing (\cite{poisson_Perez}). Figure \ref{fig:augmented_data} in the Appendix shows examples of our augmented data.

\begin{figure*}
 \centering \includegraphics[width=0.75\linewidth]{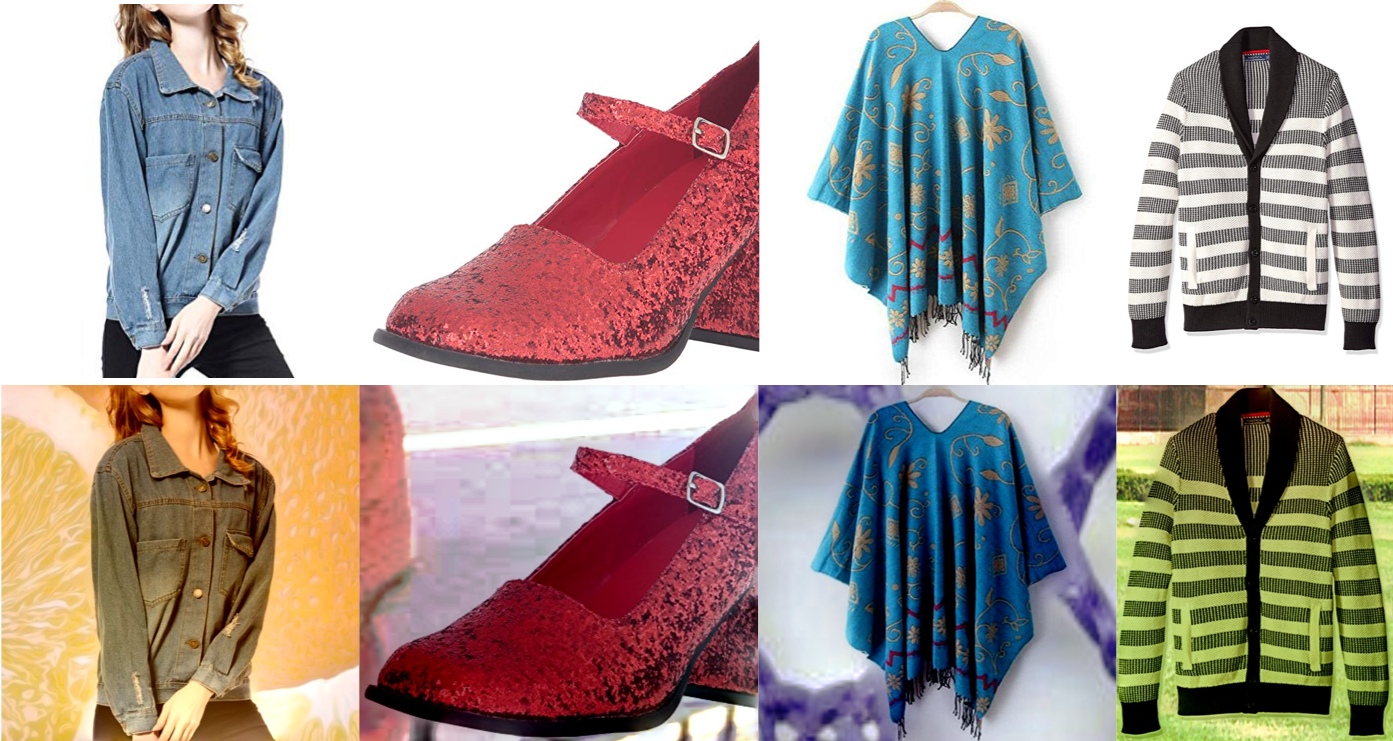}
 \caption{Augmenting catalog image with background. Note the random color blending effect across samples. Best viewed electronically}
\label{fig:augmented_data}
\end{figure*}

%\begin{figure}
%\vspace{-0.75em}
% \centering \includegraphics[width=0.4\linewidth]{Picture3.png}
% \caption{Augmenting catalog image with background. Note the random color blending effect across samples. Best viewed electronically} 
%\label{fig:augmented_data} 
%\vspace{-0.75em}
%\end{figure}

Customer review images are images that online customer uploaded when they review a product. While it is a good source for images taken in casual setting, they often are extreme close-ups and hence not useful. In addition, they are not evenly distributed. For example, there were more than 20K images for \textit{legging}, but only 1742 for \textit{denim shorts}. Nevertheless, we still sampled part of this source for our training.

%Ground truth labels for product type are directly available from the browse node tree. As for each of the attributes (\textit{color}, \textit{pattern…}), we carried out n-grams analysis on all Catalog items' title collection and identified the top frequent attributes to form attribute classification tasks. In total, we sampled 1.2M images (one image per an ASIN) and applied background augmentation to half of them. 500K of these are for product type, the rest are divided evenly among the other attributes.

For each fine-grained classifier, we use a Resnet18 that was modified for multi-task losses as described in the previous section. We initialized the network with an ImageNet pretrained model and trained it for 50 epochs using multiple rounds of cyclic learning rate (\cite{cyclic}). Using second cycle resulted in an increase of ~2\% for overall classification accuracy. (We stopped at the second round when overfitting started to appear).

\subsection*{Search Indices}

We indexed most of the apparel categories for man and woman. The main images for top items in 242 browse nodes, which correspond to 146 fine-grained classes, were selected. They were deduped according to image ids. Each image was then run through the detector (section \ref{sec:system_architecture}) to extract the bounding box for the underlying clothing item (e.g., \textit{jeans}). The cropped image patch is then fed to the corresponding fine-grained classifier network (e.g., \textit{bottom}) for embedding feature extraction. To further reduce duplication, we first placed all extracted features in a k-NN corpus, then re-queried to find near duplicate neighbors for each of the features. Based on the duplication graph, we find connected components and retained only the top Catalog item for each component.

We built one index for all Catalog items that belong to a fine-grained class. We also split indices according to gender type. In total, there are close to 500 (sub) indices. We use off-the-self \textit{hsnw} library (\cite{hnswlib/MalkovY16}) for approximate nearest neighbor search.

\section{Experimental Results}
\label{sec:experimental}
\subsection*{Localization}

In this section, we report a comparison between training detectors using our collected data vs. training using OpenImages (\cite{OpenImages}, only for clothing-related categories). It was necessary to roll-up the classes in OpenImages to make them comparable with our data, e.g. merging \textit{jean}, \textit{skirt} into \textit{bottom}. The numbers of bounding boxes are kept roughly the same on both sides (\textasciitilde 300K). We used the same detector architecture for this test (YOLOV3 416). The test set consisted of 5K held-out fashion images in the wild. When trained on OpenImages, the detector had an mAP of 0.6 on the test set, while when trained on our data (section \ref{sec:data_collection}) the mean Average Precision (mAP) \cite{Salton:1986:IMI:576628} is 0.71. This result shows the importance of dataset’s quality. We attribute the low accuracy w.r.t. OpenImages to skewness, incompleteness and inaccuracies in its bounding box annotation.
For completeness, mAP for all categories using YOLOV3 416 are listed in the Appendix. 

Using SSD 512 with a Resnet50 body, the overall mAP was slightly lower, at 0.70. We used a combination of detectors in offline processing: SSD 512 Resnet50, YOLO V3 300, YOLO V3 416. Their combination led to 2\% increase in the overall mAP.

\subsection*{Classification}

We report here the performance evaluation of fine-grained classifiers on the Catalog validation datasets without (V1) and with noise clean-up (V2) (section \ref{sec:data_collection}). We also report the classification accuracy on fashion images in the wild. We first downloaded 600K images using keywords that are corresponding to fine-grained class labels. Then we run these images through the detector ensemble in a high recall mode to extract bounding box for clothing items. All the extracted cropped images are given to Mturkers to verify their class labels. This process resulted in around 270K positive confirmation for both bounding boxes and their class labels, which we used for evaluation. In summary, the fine-grained accuracies for \textit{top} (33 fine-grained classes), \textit{bottom} (10 fine-grained classes) and \textit{dress} (5 fine-grained classes) are shown in table \ref{table:classification_accuracy}.

\begin{table*}[h!]
\begin{center}
\caption{Fine grain classification accuracies for three high-level class: \textit{top}, \textit{bottom} and \textit{dress}. *See text for further information}
\label{table:classification_accuracy}
\begin{tabular}{l*{6}{c}r}
Classifiers              & V1 - Validation & V2 - Validation & Fashion Images (270K) \\
\hline
dress  & 0.70 & 0.74 & 0.71 \\
top      & 0.64 & 0.80 & 0.52* \\
bottom  & 0.60 & 0.73 & 0.70 \\
\end{tabular}
\end{center}

\end{table*}

As can be seen from the table, cleaning up the data led to an increase of roughly 5-8\% in accuracy across the classes. It also demonstrated that although we trained our fine-grained classifiers using original and augmented Catalog images, the accuracies for \textit{dress} and \textit{bottom} only dropped slightly when tested on fashion images in the wild. For \textit{top}, there is currently an issue with the data labelling; often, the detector returns only single bounding box even if the person wears multiple layers such as a \textit{jacket} on top of a \textit{tunic}. When such a bounding box image was given to the annotator, it can be labelled as \textit{jacket} or a \textit{tunic}, while our system, by design, only predicts fine-grained class for the outwear, \textit{jacket}, in such a case. As the result, the accuracy for \textit{top} is lower than expected. We are currently working to make annotation more consistent and re-evaluate the accuracy for top on the fashion images in the wild.

\subsection*{Visual Similarity Search}

\begin{figure*}
 \centering \includegraphics[width=0.60\linewidth]{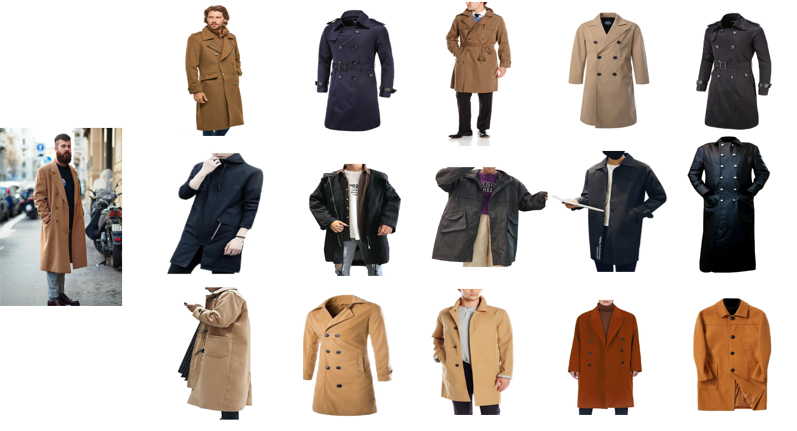}
 \caption{Left: the query image. Right: retrieval results - top row: using feature that sensitive only to product type – middle row: color aware, but trained without background augmentation – bottom row: color aware and trained with background augmentation. Best viewed electronically.} 
\label{fig:result_list} 
\end{figure*}

In this section, we show qualitative and quantitative comparisons between different versions of our feature embedding that correspond to different versions of the classification networks: (V1) single-task, \textit{product type} classification only, (V2) multi-task, \textit{product type} and \textit{color} classification, and (V3) multi-task,\textit{ product type}, \textit{color}, and additional 13 classification tasks.
In figure \ref{fig:result_list}, the picture on the left is the query image. The top row shows top-5 retrieved results using V1 embedding. Since the network only did product type classification, the resulting embedding seems to ignore other aspects such as color. The second row shows the result when a color classification task is added (V2). Initially, during training, the added task was carried out on original Catalog images only. As the result, its embedding is quite sensitive to background color which is rarely present on Catalog images. In this case, the color of the under layer was wrongly picked up for matching. However, when background augmentation was used, as shown in the third row, the feature become more resilient to background clutter.

We used humans for quantitative A/B evaluation of the retrieval quality. Exact retrieval was not targeted, rather the results were judged based on following subjective matching criteria in order of importance: occasion, product type, color, pattern followed by other clothing features. Approximately 1000 fashion images were sent out to MTurkers for evaluation. For each query image, retrieval results from two feature embeddings, A and B, were shown side by side and the Mturkers were asked to choose one of the following options: A is better than B, B is better than A, both A and B are bad and both A and B are equally good. We aggregated the votes from 5 people. Overall, between V1 and V2, people gave preference to V2 75\% of the time and to V1 25\% of the time. Between V2 and V3, the preference ratio given to V2 and V3 respectively are 33\% and 67\%. This clearly shows consistent progresses in embedding quality from V1 to V2 and from V2 to V3, as the corresponding classification network was trained to perform more tasks.

%We also did comparisons and found out that the preference ratios given to a triplet-loss based embedding (a parallel effort in our team) over V1 and V2 are 3:1 and 3:2 respectively (similar to those for V3). Comparing triplet-loss based embedding to V3 is one of our next steps. Qualitatively, we observed that while triplet-loss based embedding excelled at capturing color and pattern, classification-based embedding was better at capturing product types and attributes.
% such as \textit{sleeve length}, \textit{neck} type, etc.

\section{Conclusion and Future Directions}

\label{sec:future}
In this paper, we describe an approach of searching for apparel items using fashion images in the wild. We trained multi-box detectors to localize clothing items as well as to perform gender detection. We also trained multiple classifiers to further classify detected bounding boxes into fine-grained categories. We then used the embedding features extracted from these classifiers to search for similar clothing items from online Catalog. Our initial positive experimental outcomes validate our system design.
As the next steps, we plan to train fine-grained classifiers on fashion images in the wild (i.e. instead just on Catalog images) and to add more extensive evaluation. We also plan to use (semantic) segmentation and clothing parsing (\cite{LIP_TangsengWY17}) to improve matching performance when people wear multi-layer tops, which occur quite often in fashion images.

%\bibliographystyle{ACM-Reference-Format}
%\bibliography{sample-base}
\bibliographystyle{plain}
\bibliography{bib}{} 
 
\appendix
% \section{End to End Architecture}
% See Figure \ref{fig:overall_pipeline}
% \begin{figure*}
%  \centering \includegraphics[width=0.75\linewidth]{Picture1.png}
% \caption{Overall architecture of our end-to-end visual search for apparel} 
% \label{fig:overall_pipeline} 
% \end{figure*}

% \section{Background Augmentation}
% See Figure \ref{fig:augmented_data}
% \begin{figure*}
%  \centering \includegraphics[width=0.75\linewidth]{Picture3.png}
%  \caption{Augmenting catalog image with background. Note the random color blending effect across samples. Best viewed electronically} 
% \label{fig:augmented_data} 
% \end{figure*}
%\section{End to End Architecture}
%See Figure \ref{fig:overall_pipeline}

%\section{Background Augmentation}
%See Figure \ref{fig:augmented_data}

\section{Detector Performance}
See Table \ref{table:detection mAP}
\begin{table*}[h!]
\begin{center}
\caption{Detection mAP for one of our detectors (YOLOV3/416) on fashion images in the wild dataset}
\label{table:detection mAP}
\begin{tabular}{l*{6}{c}r}
High Level Classes              & mAP \\
\hline
Head-wear  &0.80 \\
Eye-wear	&0.81 \\
Earring	&0.49 \\
Belt	&0.53 \\
Bottom	&0.78 \\
Dress	&0.81 \\
Top	&0.86 \\
Suit	&0.67 \\
Tie	&0.67 \\
Footwear	&0.87 \\
Swimsuit	&0.54 \\
Bag	&0.79 \\
Wristwear	&0.62 \\
Necklace &0.65 \\
One-piece	 &0.60 \\
Scarf &0.52\\
\hline
Boy &0.82 \\
Girl &0.77 \\	
Woman	&0.90 \\
Man&0.81 \\
\hline
Overall      & 0.72 \\
\end{tabular}
\end{center}
\end{table*}

\end{document}